\documentclass[10pt,twocolumn,letterpaper]{article}

\usepackage[pagenumbers]{wacv} %

\usepackage[T1]{fontenc}
\usepackage[utf8]{inputenc}
\usepackage{graphicx}
\usepackage[table,xcdraw]{xcolor}
\usepackage{amsmath}
\usepackage{mathtools}
\usepackage{booktabs}
\usepackage{capt-of}
\usepackage{enumitem}

\usepackage{makecell}
\usepackage{comment}
\usepackage{amssymb}
\usepackage{bm}
\usepackage{xspace}

\usepackage{tabularx}
\usepackage{multirow}
\usepackage{microtype}

\usepackage{textcomp}
\usepackage{gensymb}

\usepackage{bbold}

\usepackage[pagebackref,breaklinks,colorlinks]{hyperref}

\usepackage{flushend}

\usepackage[accsupp]{axessibility}  %

\def\wacvPaperID{1056} %
\def\confName{WACV}
\def\confYear{2024}

\AddToShipoutPicture*{%
     \AtTextUpperLeft{%
         \put(-3.5,10){
           \begin{minipage}{\textwidth}
              \scriptsize
              \MakeUppercase{Preprint version of an IEEE/CVF Winter Conference on Applications of Computer Vision (WACV) paper; final version available at:} \url{https://ieeexplore.ieee.org}
           \end{minipage}}%
     }%
}

\begin{document}

\title{Image Labels Are All You Need for Coarse Seagrass Segmentation}

\author{Scarlett Raine$^{1,2}$, Ross Marchant$^{3}$, Brano Kusy$^{2}$, Frederic Maire$^{1}$ and Tobias Fischer$^{1}$ \\
\\
\footnotesize{$^{1}$QUT Centre for Robotics, Queensland University of Technology, Australia} {\tt \footnotesize \emph{\{sg.raine, f.maire, tobias.fischer\}@qut.edu.au}} \\
\footnotesize{$^{2}$CSIRO Data61, Australia} {\tt \emph{\{scarlett.raine, brano.kusy\}@csiro.au}} \\
\footnotesize{$^{3}$Image Analytics, Australia} {\tt \emph{ross.g.marchant@gmail.com}}%
}

\maketitle

\begin{abstract}
\vspace*{-0.2cm}

Seagrass meadows serve as critical carbon sinks, but estimating the amount of carbon they store requires knowledge of the seagrass species present. Underwater and surface vehicles equipped with machine learning algorithms can help to accurately estimate the composition and extent of seagrass meadows at scale. However, previous approaches for seagrass detection and classification have required supervision from patch-level labels.  In this paper, we reframe seagrass classification as a weakly supervised coarse segmentation problem where image-level labels are used during training (25 times fewer labels compared to patch-level labeling) and patch-level outputs are obtained at inference time. To this end, we introduce SeaFeats, an architecture that uses unsupervised contrastive pre-training and feature similarity, and SeaCLIP, a model that showcases the effectiveness of large language models as a supervisory signal in domain-specific applications. We demonstrate that an ensemble of SeaFeats and SeaCLIP leads to highly robust performance. Our method outperforms previous approaches that require patch-level labels on the multi-species `DeepSeagrass' dataset by 6.8\% (absolute) for the class-weighted F1 score, and by 12.1\% (absolute) for the seagrass presence/absence F1 score on the `Global Wetlands' dataset.  We also present two case studies for real-world deployment: outlier detection on the Global Wetlands dataset, and application of our method on imagery collected by the FloatyBoat autonomous surface vehicle.
\end{abstract}

\vspace*{-0.4cm}
\section{Introduction}
\label{sec:intro}  

\textit{Blue Carbon} refers to the capacity of coastal ecosystems to sequester significant amounts of carbon from the atmosphere~\cite{macreadie2019future}. Seagrass meadows play a vital role in this process, but the amount of carbon they store can vary depending on the seagrass species and habitat conditions~\cite{lavery2013variability}. Seagrass meadows also improve water quality, protect the coastline, and provide a source of food and nursery shelter for fish~\cite{maxwell2019seagrasses, raine2020multi}. Scientists require detailed data on seagrass meadow extent and composition to accurately estimate blue carbon sequestration~\cite{lavery2013variability}, as well as for supporting the long-term ecosystem management and conservation of these critical ecosystems~\cite{macreadie2019future}.

\begin{figure}[t]
\centering
\includegraphics[width=0.98\linewidth,clip, trim=2.95cm 1.2cm 4.85cm 2.35cm]{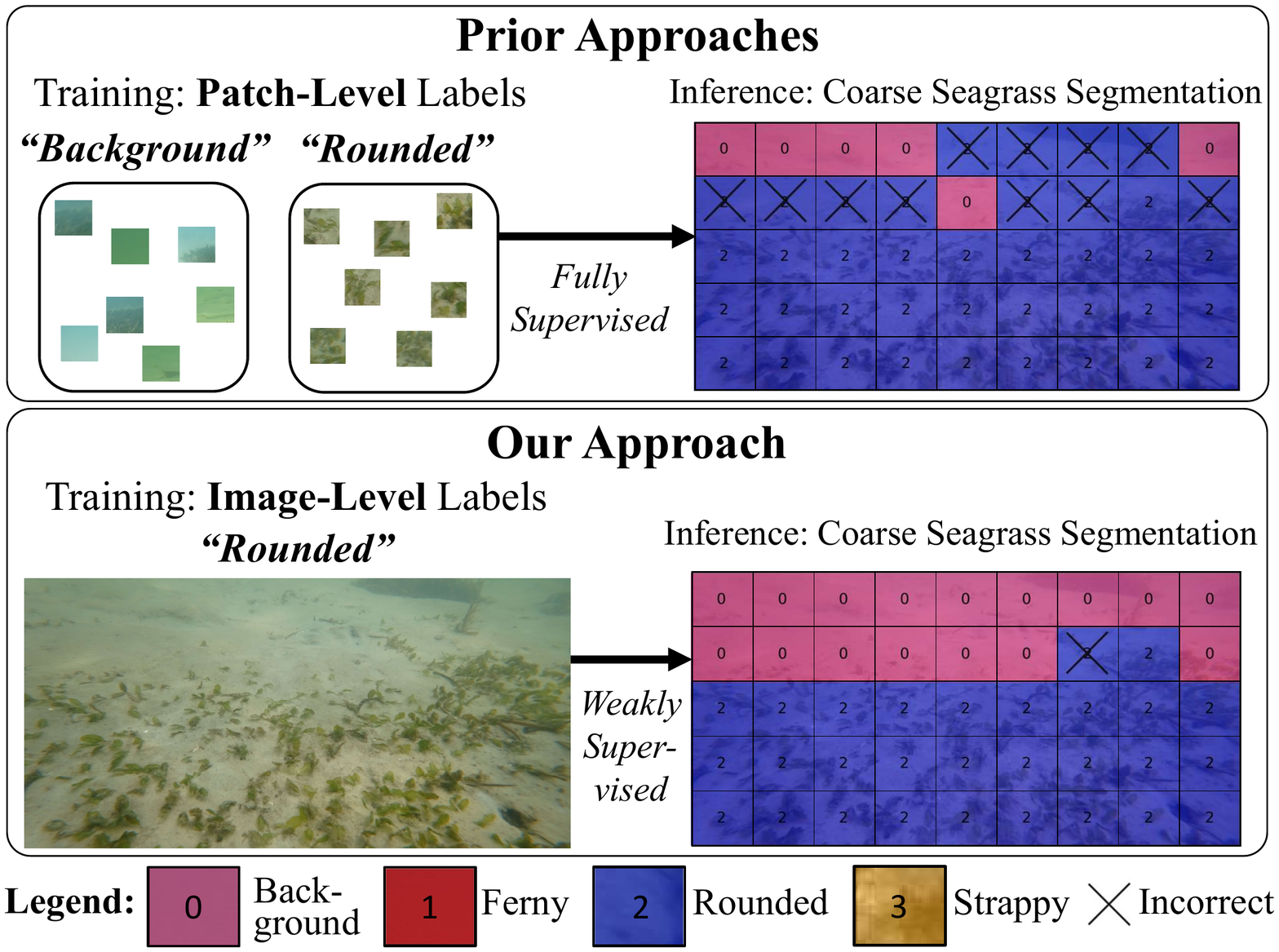}
\vspace*{-0.6cm}
\caption{Top: Prior approaches for multi-species seagrass detection and classification relied on fully-supervised training with patch-level labels.  Bottom: Our proposed approach is weakly supervised, requiring only image-level labels, while still achieving patch-level coarse seagrass segmentation at inference time. Refer to Section \ref{subsec:datasets} for definition of classes `Background', `Ferny', `Rounded' and `Strappy'.}
\vspace*{-0.7\baselineskip}
\label{fig:frontpage}
\end{figure}

Autonomous and semi-autonomous underwater survey methods are critical for large-scale underwater imaging \cite{li2022real, kusy2022situ, mou2022reconfigurable} and habitat surveys~\cite{ferrari2018large}.  Underwater vehicles can validate remote geospatial and aerial sensing technology, analyze seagrass species distributions after disasters, and monitor climate change \cite{maxwell2019seagrasses}. 

Machine learning enables automated, accurate processing of image data and allows ecologists to adapt survey transect paths in real-time \cite{li2022real, koreitem2020one}.  However, deep learning approaches require large amounts of labeled data for supervised training, but underwater image pixel-level labeling by domain experts is costly and time-consuming \cite{modasshir2020enhancing}. Seagrass also lacks distinct semantic features and has poorly defined boundaries, making it challenging to detect using typical object detection methods. An alternative to pixel-level labeling is grid-based patch labeling, which reduces annotation costs while enabling seagrass detection and classification~\cite{raine2020multi, noman2021multi}. This is consistent with approaches for automated benthic habitat classification \cite{chen2021new}, which typically employ grid-based or point-grid labeling styles. We refer to class labels assigned to image patches as patch-level labels, while those assigned to the entire image are image-level labels (Fig.~\ref{fig:frontpage}).

Earlier work in multi-species seagrass classification naively assigned each image patch the image-level label~\cite{raine2020multi}, which facilitates fast training but requires images that only contain seagrass -- any regions of sand, water, or other objects introduce noise into the dataset.
Other research explored a teacher-student framework to enable fast training on unlabeled image patches, however their teacher model requires patch-level labels~\cite{noman2021multi}.

In this paper, we propose a principled approach to coarse segmentation from image-level labels. Our framework is an ensemble of two complementary classifiers which leverage unsupervised pre-training and the general semantic knowledge of a large vision-language model to propose pseudo-labels during training.

\noindent Our paper presents the following key contributions:
\begin{enumerate}[topsep=0pt,itemsep=-1ex,partopsep=1ex,parsep=1ex]
    \item We re-frame seagrass classification as a weakly supervised coarse segmentation task, such that only image-level labels are required for training, but patch-level classifications are obtained at inference time.  
    \item We present an ensemble of two novel methods for multi-species coarse segmentation: `SeaFeats' uses a novel loss function and classifies patches into background and seagrass by comparing patch features to class templates.  `SeaCLIP' demonstrates that the CLIP large language model \cite{radford2021learning} can be effectively used as a supervisory signal in domain-specific applications.
    \item We present exhaustive experimental trials, in which we outperform the state-of-the-art by 6.8\% (absolute F1 score) on the DeepSeagrass dataset \cite{raine2020multi}. We also provide a variety of ablations that investigate the effectiveness of the various components in our model. 
    \item We perform two real-world deployment case studies:  First, we contribute a labeled test set for the `Global Wetlands' dataset \cite{ditria2021avff} and demonstrate that SeaCLIP can be effectively used for outlier detection of fish.  Second, given just 20 labeled images, we demonstrate generalization capability to underwater imagery collected by the FloatyBoat \cite{mou2022reconfigurable} autonomous surface vehicle.
\end{enumerate}

\noindent We make our code available to foster future research on coarse seagrass segmentation\footnote{\url{https://github.com/sgraine/bag-of-seagrass}}.
\section{Related Work}
\label{sec:related}

Research on coral image classification has received considerable attention from researchers~\cite{gonzalez2020monitoring, modasshir2018mdnet, gomez2019coral, koreitem2020one, raine2022point, modasshir2020enhancing}. In contrast, seagrass detection and classification has not been explored as much \cite{reus2018looking, weidmann2019closer, pamungkas2021segmentation, lestari2021segmentation, noman2021seagrass}, with only a few studies attempting multi-species segmentation~\cite{tahara2022species, roman2021using, hobley2021semi,raine2020multi, noman2021multi}. We detail these multi-species approaches in Section~\ref{subsec:multi-seagrass}. To address the problem of multi-species seagrass detection and classification, we frame it as a weakly supervised problem, and explore existing methods in this area in Section~\ref{subsec:weakly}. In addition, we discuss recent advances in large vision-language models and their relevance for weak supervision and outlier detection in the field of marine surveys in Section~\ref{subsec:clip-lit}.

\subsection{Multi-species Seagrass Detection/Classification}
\label{subsec:multi-seagrass}

In the context of multi-species mapping of seagrass at scale, numerous methods have been developed using data collected via unmanned aerial vehicles \cite{tahara2022species, roman2021using} or remotely piloted aircraft \cite{hobley2021semi}. However, these remote approaches have limitations in terms of their ability to discriminate between seagrass species, and can only operate effectively during low wind conditions, at low tide and at certain times of the day \cite{tahara2022species, roman2021using}. %

In contrast, Raine \etal \cite{raine2020multi} posed the problem of in-situ multi-species seagrass classification of underwater images. Specifically, \cite{raine2020multi} contributed the \textit{`DeepSeagrass'} dataset containing seagrass image patches taken from the viewpoint of an underwater vehicle, and presented a method where each patch is naively assigned the label of the parent image. Noman \etal \cite{noman2021multi} used pseudo ground-truth labels generated by a teacher model to train a student model.  Noman \etal \cite{noman2021multi} also contributed the \textit{`ECU-MSS'} multi-species seagrass dataset, which unfortunately is not publicly available\footnote{We contacted the authors, who are not able to share the dataset.} and therefore could not be used for evaluation of our methods.

To address the limitations of these approaches, we propose a structured framework for image-level weak supervision for coarse seagrass segmentation, without naively assuming the patch-level labels or requiring time-consuming patch-level labeling by domain experts.

\begin{figure*}[ht]
\vspace*{0.2cm}
\centering
\includegraphics[width=0.97\textwidth, clip, trim=1.8cm 4.75cm 1.5cm 4.75cm]{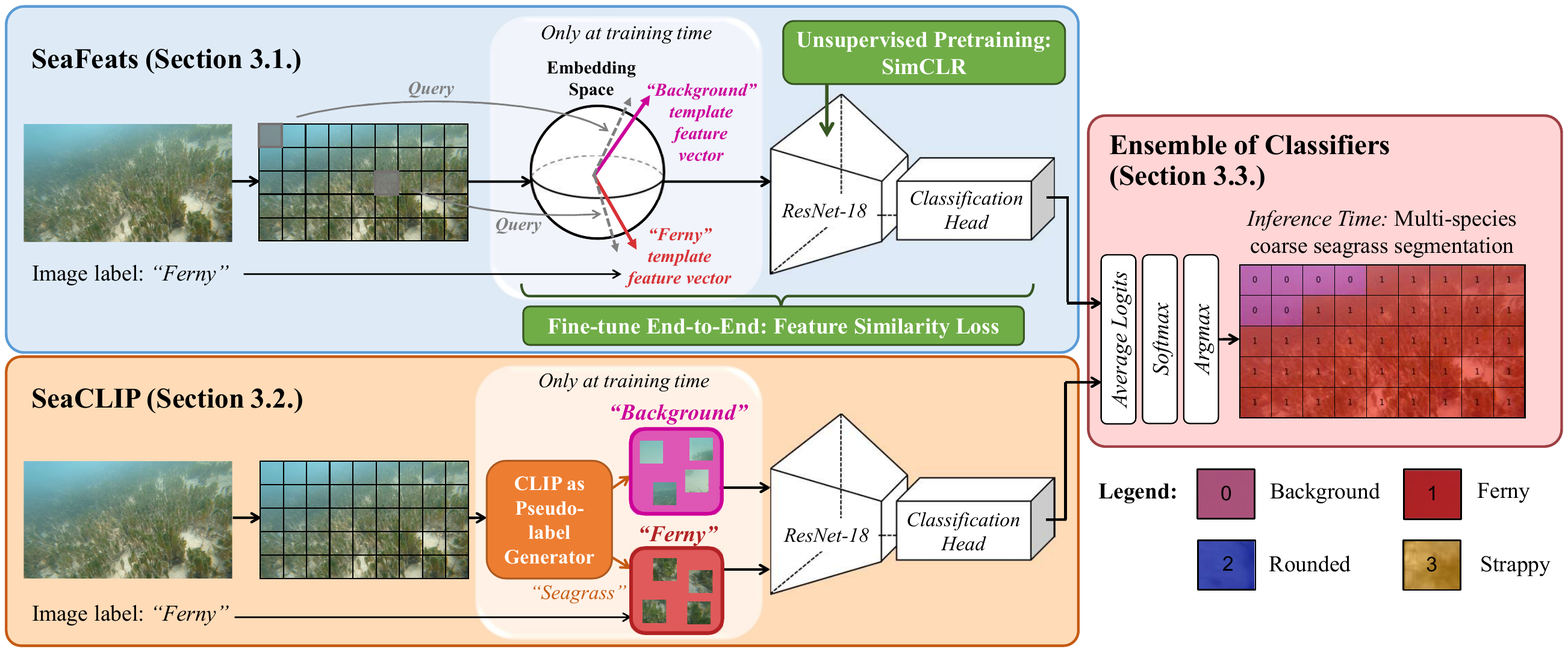}
\vspace*{-0.3cm}
\caption{Proposed Algorithm Schematic. We propose an ensemble of two complementary methods for coarse seagrass segmentation.  Top: SeaFeats is trained by fine-tuning a feature extractor (initialized with weights from an unsupervised contrastive task) and classifier by finding the cosine similarity between patch-level feature vectors and our per-class template feature vectors (Section \ref{subsec:met-seafeats}).  Bottom: SeaCLIP is trained using zero-shot pseudo-labels predicted by the pre-trained large language model CLIP \cite{radford2021learning}, in combination with the image-level domain-specific seagrass labels (Section \ref{subsec:met-seaclip}).  Right: At inference time, SeaFeats and SeaCLIP are combined in parallel as an ensemble of classifiers and predict the coarse segmentation mask of the query image (Section \ref{subsec:met-ensemble}).}
\label{fig:pipeline}
\vspace*{-0.3cm}
\end{figure*}

\subsection{Weakly Supervised Object Localization}
\label{subsec:weakly}
Pixel-level and patch-level labeling of underwater imagery by domain experts is prohibitively costly and time-consuming. However, underwater recognition is dominated by methods which rely on expert knowledge and human labeling \cite{gonzalez2023survey}.  To address this challenge, we explore weakly supervised object detection and localization, where the model is trained only on the image-level label to determine the location and class of instances in images \cite{cole2022label, shao2022deep, zhang2021weakly}. However, our setting is unique in that we focus on coarse seagrass segmentation. Seagrass grows as a `carpet' along the seafloor, meaning that all regions of the inference image must be classified, and bounding box proposal is ineffective as there is no clearly defined object/background boundary.

Li \etal \cite{li2021dual} performed medical whole slide image classification and patch-level tumor localization via self-supervised contrastive pre-training -- we adapt such pre-training to distinguish different types of seagrass. 

To the best of our knowledge, coarse seagrass segmentation in underwater imagery from image-level labels has not been attempted previously.  To this end, we formulate this problem as a weakly supervised task, and propose our novel ensemble of classifiers as an innovative and effective solution.

\subsection{Vision-Language Models}
\label{subsec:clip-lit}
In recent years, large vision-language models have garnered significant interest for their potential to perform zero-shot classification of images.   These models typically consist of two encoders: one for textual descriptions and another for visual representations \cite{jia2021scaling, li2021align, radford2021learning}.  By jointly training these encoders, the models learn to map the textual descriptors and image-based features into a shared embedding space.  One notable large vision-language model, CLIP for `Contrastive Language-Image Pre-training', was trained on 400 million image-caption pairs collected from the internet \cite{radford2021learning}. The substantial amount of training data, combined with the application of contrastive learning to establish a shared embedding space, results in CLIP possessing a general understanding of visual and semantic concepts, making it applicable to unseen domains and capable of being queried with natural language descriptions of novel classes \cite{radford2021learning}.  

Many recent approaches build on CLIP for tasks including zero-shot out-of-distribution detection \cite{esmaeilpour2022zero}, referring image segmentation \cite{wang2022cris, luddecke2022image} and zero-shot segmentation \cite{zhou2022zegclip, li2022languagedriven}. However, using CLIP as a training signal for patch-level detection and classification in underwater images has not yet been attempted.  Furthermore, the general knowledge embedded in the CLIP model has not been leveraged in conjunction with weak, domain-specific annotations.  We believe that introducing the use of large vision-language models to the fields of ecology and marine surveys is a novel and important contribution.

\section{Method}
\label{sec:method}

We propose two novel approaches for coarse seagrass segmentation from image-level labels: SeaFeats and SeaCLIP. To eliminate the supervision required from patch-level annotations, both methods exhibit internal mechanisms for proposing pseudo-labels within training images.  Firstly, SeaFeats uses the similarity of deep features to determine whether patches are closer in the embedding space to the background or seagrass species template feature vectors (Section~\ref{subsec:met-seafeats}).  Secondly, SeaCLIP demonstrates that the CLIP vision-language model can be used as a supervisory signal for proposing which patches contain seagrass \mbox{(Section~\ref{subsec:met-seaclip})}.  By querying CLIP with text phrases describing seagrass, SeaCLIP is particularly effective for blurry, distant or image edge patches -- in these cases, it behaves conservatively and will classify patches as background. We therefore combine our two methods in parallel as an ensemble of classifiers, and find that this improves the robustness of our method (Section~\ref{subsec:met-ensemble}). Our full pipeline is shown in Fig.~\ref{fig:pipeline}.  

At inference time, only the ResNet-18 classifiers are required, enabling light-weight deployment and comparable inference times to the state-of-the-art~\cite{raine2020multi,noman2021multi}.

\subsection{SeaFeats -- Feature Similarity}
\label{subsec:met-seafeats}

Inspired by hyperdimensional computing \cite{neubert2021hyperdimensional, wilson2023hyperdimensional, neubert2021vector}, SeaFeats proposes pseudo-labels by generating high-dimensional template feature vectors for each of the classes (background and each of the seagrass species), and then measuring the similarity of each image patch to these template vectors (Fig.~\ref{fig:pipeline}, top; Eq.~(\ref{eq:templates})). Given these pseudo-labels, we then train a coarse segmentation framework based on a ResNet-18 encoder followed by a classification head.  We fine-tune the architecture end-to-end using our custom feature similarity loss function (Section~\ref{subsubsec:loss}).  We initialize the encoder with weights learned using an unsupervised pretext task (Section~\ref{subsubsec:simclr}). At inference time, we take the trained SeaFeats ResNet-18 encoder and classification head and directly predict the output coarse segmentation mask of a query image. 

\subsubsection{Feature Similarity Loss Function}
\label{subsubsec:loss}
We propose a novel loss function inspired by feature aggregation and template matching in hyperdimensional computing \cite{neubert2021hyperdimensional, wilson2023hyperdimensional, neubert2021vector}.  After every epoch, we extract feature vectors from the final layer of the ResNet-18 encoder for each of the patches.  We obtain a template feature vector for each class by averaging the L2 normalized patch features corresponding to each class at the image level: 
\begin{equation}
    \label{eq:templates}
        \bar{\textbf{v}}_c = \frac{1}{N_c}  \sum_x \mathbb{1}_{[x=c]} \frac{ \textbf{v}_x }{ ||\textbf{v}_x||}_{2},
\end{equation}
where $\textbf{v}_x$ is the extracted feature vector for patch $x$ and $\mathbb{1}_{[x=c]}$ is an indicator function which returns 1 if the image label $x$ of patch $\textbf{v}_x$ is equal to class $c$, and 0 otherwise.  Therefore, for each class $c$, we compute $ \bar{\textbf{v}}_c$, the average of the normalized feature vectors from a sample of $N_c$ patches from class $c$.  It is important to note that the per-class template feature vectors are dynamically updated every epoch as the feature extractor is fine-tuned during training.

During training, we calculate the cosine similarity between our current patch's feature vector $\textbf{v}_x$ and the template feature vector $\bar{\textbf{v}}_c$ corresponding to the image-level label $c$:
\begin{equation}
    \label{eq:cosine}
        \mathrm{sim(}\textbf{v}_x, \bar{\textbf{v}}_c) = \frac{ \textbf{v}_x \cdot \bar{\textbf{v}}_c }{ ||\textbf{v}_x|| \;||\bar{\textbf{v}}_c|| }.
\end{equation}

We also compute the cosine similarity between $\textbf{v}_x$ and the background template feature vector $\bar{\textbf{v}}_0$. If the similarity to the background template feature vector is higher, then the patch will be pseudo-labeled as background, despite the seagrass image-level label. Otherwise, the pseudo-label for the patch will correspond to the image-level label,
\begin{equation}
p_x = \begin{cases} 
          0 & \mathrm{sim(}\textbf{v}_x, \bar{\textbf{v}}_0) > \mathrm{sim(}\textbf{v}_x, \bar{\textbf{v}}_c)  \\
          c & \mathrm{otherwise}
       \end{cases}
\end{equation}
where $p_x$ is the pseudo-label for patch $x$ and the \mbox{class index $0$} corresponds to the background class.  We use the categorical cross entropy between the model's output softmax values and the pseudo-label to train the neural network. 

\subsubsection{Contrastive Pretext Task}
\label{subsubsec:simclr}
As our loss function relies on features extracted from the encoder of our architecture, the initialization of the feature encoder is important for high performance \cite{li2021dual}.  We use an unsupervised pretext task to obtain a pre-trained ResNet-18 feature extractor, as in \cite{li2021dual}.  We add our classification head (Section~\ref{subsubsec:exp-seafeats}) and then fine-tune the architecture end-to-end with the loss function described in Section~\ref{subsubsec:loss}.  

Specifically, the pretext task we consider uses the SimCLR \cite{chen2020simple} contrastive learning framework. For a batch of $B$ images, SimCLR creates two random `views' for each image using data augmentation transforms, giving a total of $2B$ images per batch (see Section~\ref{subsubsec:exp-seafeats})  \cite{chen2020simple}. For each image, the corresponding two augmented views are considered as a `positive pair', while all other samples in the batch are negative samples \cite{chen2020simple}. The network is trained to maximize the cosine similarity between the representation of a sample, $\textbf{v}_i$, and its positive view $\textbf{v}_j$, and minimize the similarity to the negative samples, $\{\textbf{v}_k\}_{k\in\{1,...,2B\}, k \neq i}$, using the following contrastive loss~\cite{chen2020simple}:  
\begin{equation}
    l_{i,j} = -\log \frac{ \mathrm{exp(sim(}\textbf{v}_i,\textbf{v}_j)/\tau)}{\sum_{k=1}^{2B} \mathbb{1}_{[k\neq i]} \mathrm{exp(sim(}\textbf{v}_i,\textbf{v}_k)/ \tau)},
\end{equation}
where $\tau$ is a temperature parameter which we set to $\tau=0.07$ as recommended.

\begin{comment}
We use the SimCLR \cite{chen2020simple} contrastive learning framework to train our encoder as a pretext task.  SimCLR creates random batches of samples where the first sample in the batch is heavily augmented (see Section~\ref{subsubsec:exp-seafeats})  \cite{chen2020simple}.  The original sample and the augmented sample are considered as the `positive pair', while all other samples in the batch are negative samples \cite{chen2020simple}. The following contrastive loss \cite{chen2020simple} is used to identify the positive pair $(i,j)$ in the batch:
\begin{equation}
    l_{i,j} = -\log \frac{ \mathrm{exp(sim(}\textbf{v}_i,\textbf{v}_j)/\tau)}{\sum_{k=1}^{2B} \mathbb{1}_{[k\neq i]} \mathrm{exp(sim(}\textbf{v}_i,\textbf{v}_k)/ \tau)},
\end{equation}
%
where $\mathrm{sim(}\textbf{v}_i,\textbf{v}_j)$ is the cosine similarity between deep features for samples $(i,j)$, $\mathbb{1}_{[k\neq i]} \in \{0,1\}$ is an indicator function which returns $1$ if $k \neq i$ and $0$ otherwise, $B$ is the number of original samples in the batch, and $\tau$ is a temperature parameter which we set to $\tau=0.07$ as recommended.
\end{comment}

%
\subsection{SeaCLIP -- Large Vision-Language Model}
\label{subsec:met-seaclip}
 
We leverage the pre-trained large vision-language model, CLIP \cite{radford2021learning} as a zero-shot pseudo-label generator. CLIP has been trained jointly on image and text embeddings.  While CLIP has extremely limited specialist knowledge in seagrass species, it has general semantic knowledge of sand, water and seagrass, which we leverage to predict which patches in training images contain seagrass and which should be labeled as background. Note that CLIP by itself cannot be used to distinguish between different seagrass species. 

We use the original CLIP pre-trained model and follow the recommended method for zero-shot classification \cite{radford2021learning}. As CLIP is trained on natural language prompts, it can be leveraged in this way to filter out any out-of-distribution objects or noisy image patches.  We obtain binary background / generic seagrass pseudo-labels from CLIP (see Section~\ref{subsubsec:exp-seaclip} and the Supplementary Material for further details on the prompts we use) and assign the single image-level species-specific seagrass label to the patches pseudo-labeled as generic seagrass by CLIP. We then train our SeaCLIP classifier (Section \ref{subsubsec:exp-seaclip}) end-to-end.  In this way, SeaCLIP combines CLIP's general semantic knowledge with domain-specific expertise to obtain accurate patch-level pseudo-labels for training. At inference time, the trained SeaCLIP ResNet-18 encoder and classification head are used directly to predict the class of all query image patches, without needing the expensive inference of CLIP.

\subsection{Ensemble of Classifiers}
\label{subsec:met-ensemble}

We find that our two proposed methods, SeaFeats and SeaCLIP, are complementary classifiers that result in improved performance when combined in an ensemble (Fig.~\ref{fig:pipeline}).  For example, we observe that SeaCLIP is more likely to classify visually degraded patches as background, rather than over-confidently misclassifying the species. When combined with SeaFeats, this is a desirable property as it prevents incorrect species identification of visually degraded patches, resulting in improved overall robustness of our method.  We integrate the two classifiers by averaging the normalized output logits and then applying the softmax function.  We find that combining the classifier outputs prior to applying softmax preserves the magnitudes of the logits, thereby resulting in more confident predictions \cite{tassi2022impact}. The scale of the logits from each classifier is normalized prior to averaging to ensure that both classifiers are `weighted' equally in this operation. 
\section{Experimental Setup}
\label{sec:experimentalsetup}
In this section, we briefly discuss implementation details (Section~\ref{subsec:Implementation}), evaluation datasets (Section~\ref{subsec:datasets}), and evaluation metrics used (Section~\ref{subsec:evaluationmetrics}).

\begin{table*}[ht]
\setlength{\tabcolsep}{5.5px}
\vspace*{0.15cm}
\caption{Multi-Species Seagrass Classification Results -- DeepSeagrass Dataset (Refer to Section~\ref{subsec:evaluationmetrics} for Metric Definitions)} 
\vspace*{-0.15cm}
\label{table:multi-results}
\centering
\scriptsize
\begin{tabular}{@{} lccccccccccccccc @{}} %
 \toprule %
\multirow{2}[3]{*}{\textbf{Method}} & \multirow{2}[3]{*}{\textbf{Labels}} & \multicolumn{3}{c}{\textbf{Background}} & \multicolumn{3}{c}{\textbf{Ferny}} & \multicolumn{3}{c}{\textbf{Rounded}} & \multicolumn{3}{c}{\textbf{Strappy}} & \textbf{Overall} & \textbf{Inference}\\
 \cmidrule(lr){3-5} \cmidrule(lr){6-8} \cmidrule(lr){9-11} \cmidrule(lr){12-14}
 & & Prec. & Recall & F1 & Prec. & Recall & F1 & Prec. & Recall & F1 & Prec. & Recall & F1 & \textbf{F1 Score} & \textbf{Time (s)} \\
\midrule
Zero-shot CLIP \cite{radford2021learning} & Nil & 79.92 & 87.23 & 83.41 & 86.26 & 20.55 & 33.20 & 56.35 & 34.75 & 42.99 & 36.62 & 82.72 & 50.76 & 60.65 & 0.783 \\
\arrayrulecolor{black!30}\midrule
SimCLR \cite{chen2020simple}  & \multirow{2}{*}{Patch} & \multirow{2}{*}{99.64} & \multirow{2}{*}{83.45} & \multirow{2}{*}{90.83} & \multirow{2}{*}{84.63} & \multirow{2}{*}{53.70} & \multirow{2}{*}{65.70} & \multirow{2}{*}{51.24} & \multirow{2}{*}{96.58} & \multirow{2}{*}{66.96} & \multirow{2}{*}{60.90} & \multirow{2}{*}{93.41} & \multirow{2}{*}{73.73} & \multirow{2}{*}{77.16} & \multirow{2}{*}{\textbf{0.019}} \\
+ Raine \etal \cite{raine2020multi} \\
\midrule
Raine \etal  & \multirow{2}{*}{Patch} & \multirow{2}{*}{99.76} & \multirow{2}{*}{70.97} & \multirow{2}{*}{82.94} & \multirow{2}{*}{79.45} & \multirow{2}{*}{\textbf{99.47}} & \multirow{2}{*}{88.34} & \multirow{2}{*}{86.10} & \multirow{2}{*}{97.58} & \multirow{2}{*}{91.48} & \multirow{2}{*}{86.01} & \multirow{2}{*}{97.87} & \multirow{2}{*}{91.56} & \multirow{2}{*}{87.41} & \multirow{2}{*}{0.022} \\
ResNet-50 \cite{raine2020multi} \\
\midrule
Noman \etal  & \multirow{2}{*}{Patch} & \multirow{2}{*}{\textbf{99.90}} & \multirow{2}{*}{73.11} & \multirow{2}{*}{84.43} & \multirow{2}{*}{88.00} & \multirow{2}{*}{99.01} & \multirow{2}{*}{93.18} & \multirow{2}{*}{86.56} & \multirow{2}{*}{\textbf{99.33}} & \multirow{2}{*}{92.50} & \multirow{2}{*}{76.26} & \multirow{2}{*}{\textbf{99.04}} & \multirow{2}{*}{86.17} & \multirow{2}{*}{88.52} & \multirow{2}{*}{0.047} \\
EfficientNet-B5 \cite{noman2021multi} \\
\midrule
Ours:   & \multirow{2}{*}{Image} & \multirow{2}{*}{97.47} & \multirow{2}{*}{\textbf{92.59}} & \multirow{2}{*}{\textbf{94.97}} & \multirow{2}{*}{\textbf{92.19}} & \multirow{2}{*}{98.89} & \multirow{2}{*}{\textbf{95.42}} & \multirow{2}{*}{\textbf{93.50}} & \multirow{2}{*}{95.92} & \multirow{2}{*}{\textbf{94.69}} & \multirow{2}{*}{\textbf{97.57}} & \multirow{2}{*}{95.06} & \multirow{2}{*}{\textbf{96.29}} & \multirow{2}{*}{\textbf{95.33}} & \multirow{2}{*}{0.025}\\
SeaFeats+SeaCLIP \\
\arrayrulecolor{black!100}\bottomrule 
\end{tabular}
\vspace*{-0.25cm}
\end{table*}

\subsection{Implementation}
\label{subsec:Implementation}
All experiments are conducted with an NVIDIA A100 GPU. Both classifiers were implemented in PyTorch \cite{paszke2019pytorch}. In the following, we discuss the hyperparameters and implementation details of our approach. 

\vspace*{-0.1cm}
\subsubsection{SeaFeats -- Feature Similarity}
\label{subsubsec:exp-seafeats}
We design our architecture as a ResNet-18 \cite{he2016deep} encoder with the final fully connected layers removed, followed by an average pooling layer which applies the classification head to patches in an inference image.  Following \cite{raine2020multi}, our classification head is comprised of a fully connected layer with 512 nodes and ReLU activation, dropout with probability of 0.15 to prevent over-fitting, and a final fully connected layer with one node per class. We train our network with the Adam optimizer with an initial learning rate of 0.00001, a batch size of 3 images and a maximum of 150 epochs.  To mitigate class imbalance, we weight the categorical cross entropy with per-class weights of $ w = [1.0,1.5,1.2,1.2]$ for classes $[$`Background', `Ferny', `Rounded', `Strappy'$]$.

We initialize our encoder with weights obtained from unsupervised contrastive pre-training. To this end, we use the SimCLR implementation at \cite{thalles_silva_2021_4486327}. We feed patches of resolution 520x578, and set the random cropping operation to 132x132 pixels.  We use 132 patches per batch, train for 200 epochs on 42,848 unlabeled training patches from the DeepSeagrass dataset, and use resizing and cropping, horizontal flipping, vertical flipping, color jitter and Gaussian blur to create the positive samples during training. 

\vspace*{-0.1cm}
\subsubsection{SeaCLIP -- Large Vision-Language Model}
\label{subsubsec:exp-seaclip}
SeaCLIP leverages the pre-trained and publicly available CLIP model reported in \cite{radford2021learning}. To improve the zero-shot performance of CLIP as a pseudo-label generator, we employ basic prompt engineering, where we design multiple query prompts in various forms, \eg ``a photo of seagrass'', ``a blurry photo containing some seagrass'' or ``a photo of grass-like leaves underwater'' (refer to the Supplementary Material for further details).  If any one of these prompts has the highest associated probability, then the patch would be pseudo-labeled as the image-level seagrass species in our method. Similarly to our SeaFeats model, we train a ResNet-18 \cite{he2016deep} architecture initialized with ImageNet weights using weighted categorical cross entropy and the Adam optimizer with a batch size of 32 patches. 
 
\subsection{Datasets}
\label{subsec:datasets}

The \textbf{DeepSeagrass} dataset \cite{raine2020multi} consists of 1,701 high resolution (4624x2600 pixels) training images.  Eight geographic areas that are not overlapping with the training images were reserved for a test dataset containing 335 images \cite{raine2020multi}.  Images were divided into 40 (8x5) smaller patches based on the image-level class.  Prior to performing trials, we manually validated the test dataset and out of the 13,378 test patches, we corrected the class of 591 patches and removed 163 ambiguous patches.  

DeepSeagrass considers four broad classes: `Background', consisting of patches with less than 1\% total seagrass, \eg water column, sand or regions of the image which are too blurry to distinguish; `Ferny' (\textit{Halophila spinulosa}); `Rounded' (\textit{Halophila ovalis}); and `Strappy' (\textit{Cymodocea serrulata}, \textit{Halodule uninervis}, \textit{Syringodium isoetifolium} and \textit{Zostera muelleri})~\cite{raine2020multi}. 

We propose to use the \textbf{Global Wetlands} luderick and seagrass dataset \cite{ditria2021avff} to consider the scenario of a robot deployment of our methods.  When an ecologist collects a new set of training images, they will likely contain out-of-distribution objects which the ecologist is not interested in. Global Wetlands contains three broad classes: seagrass, background and fish (out-of-distribution).  We also use Global Wetlands to evaluate the domain generalization of our approach and for validating our CLIP-supervised method.  

Global Wetlands was collected in Moreton Bay, Australia, using remote underwater video cameras. For evaluation, we use the 764 `novel-test' split images which were collected from a different geographic location than the `train' and `test' splits. We create 50 (10x5) patches from each image, resulting in 38,200 test image patches, and manually label the patches into `fish', `seagrass' and `background' (refer to the Supplementary Material for further details). The seagrass in this dataset corresponds to the `Strappy' seagrass morphotype in the DeepSeagrass dataset. We publicly release the labeled test set to facilitate future evaluation on this dataset\footnote{\url{https://doi.org/10.5281/zenodo.7659203}}.

\vspace*{-0.1cm}
\subsubsection{FloatyBoat}
\label{subsubsec:data-floatyboat}
We also consider the deployment of our methods onboard an autonomous surface vehicle and evaluate the domain generalization of our approach on underwater imagery collected by the FloatyBoat autonomous surface vehicle \cite{mou2022reconfigurable} at Lizard Island, Australia.  This imagery exhibits a profound domain shift from the DeepSeagrass dataset, with different camera properties, viewpoint (DeepSeagrass is taken from the oblique angle at 45\degree, while the FloatyBoat camera is vertically top-down), lighting, turbidity, geographical location and seagrass stage of growth and density (DeepSeagrass contains dense seagrass with $>$70\% cover, while the FloatyBoat seagrass meadow is sparse). We utilize a survey transect of 1,148 images to evaluate our approach. Prior to applying our method, we perform basic automatic color correction.  

\subsection{Evaluation Metrics}
\label{subsec:evaluationmetrics}

We report the performance of our method using the per-class precision, recall, F1 scores, and the overall F1 score.  We predominantly use the F1 scores because they balance the importance of precision and recall:
\begin{equation}
    \mathrm{F_1} = 2 \frac{ \mathrm{precision} \cdot \mathrm{recall}}{ \mathrm{precision}+\mathrm{recall} }.
    \label{eq:f1}
\end{equation}

We note that prior approaches have been tuned for high seagrass recall values, however this is at the expense of precision, \ie many background patches are incorrectly predicted as seagrass.  In the context of seagrass density estimation, it is important to accurately predict the presence of seagrass to prevent overestimation of coverage and carbon stock, and this is better captured by the F1 scores.  

We evaluate the performance of our method in the multi-species use case, in which we consider the four categories described in Section~\ref{subsec:datasets}, as well as the binary (background/seagrass) setting for estimation of seagrass presence/absence.

\begin{table}[t]
\setlength{\tabcolsep}{4px}
\vspace*{0.15cm}
\caption{Binary Seagrass Classification -- DeepSeagrass F1 Scores (Refer to Section~\ref{subsec:evaluationmetrics} for Metric Definitions)} 
\vspace*{-0.2cm}
\label{table:binary-results}
\centering
\scriptsize
\begin{tabularx}{\columnwidth}{@{}l>{\centering\arraybackslash}X>{\centering\arraybackslash}X>{\centering\arraybackslash}X>{\centering\arraybackslash}X}
\toprule
\textbf{Method} & \textbf{Labels} & \textbf{Background} & \textbf{Seagrass} & \textbf{Overall}\\
\midrule
Zero-shot CLIP \cite{radford2021learning} & Nil & 83.41 & 87.74 & 85.90 \\
\arrayrulecolor{black!30}\midrule
SimCLR \cite{chen2020simple} & \multirow{2}{*}{Patch} & \multirow{2}{*}{90.85} & \multirow{2}{*}{94.55} & \multirow{2}{*}{93.17} \\
+ Raine \etal \cite{raine2020multi} \\
\midrule
Raine \etal & \multirow{2}{*}{Patch} & \multirow{2}{*}{82.94} & \multirow{2}{*}{90.90} & \multirow{2}{*}{88.13} \\
ResNet-50 \cite{raine2020multi} \\
\midrule
Noman \etal & \multirow{2}{*}{Patch} & \multirow{2}{*}{84.42} & \multirow{2}{*}{91.54} & \multirow{2}{*}{89.03} \\
EfficientNet-B5 \cite{noman2021multi} \\
\midrule
Ours: & \multirow{2}{*}{Image} & \multirow{2}{*}{\textbf{95.01}} & \multirow{2}{*}{\textbf{96.72}} & \multirow{2}{*}{\textbf{96.04}} \\
SeaFeats+SeaCLIP \\
\arrayrulecolor{black!100}\bottomrule 
\end{tabularx}
\vspace*{-0.2cm}

\end{table}

\section{Results}
In Section~\ref{subsec:sota}, we compare our method to the multi-seagrass detection and classification state-of-the-art~\cite{raine2020multi,noman2021multi} and also compare our approach to two widely-used classification methods, SimCLR \cite{chen2020simple} and CLIP \cite{radford2021learning}. In Section~\ref{subsec:ablations}, we perform ablation studies to demonstrate the effect of feature extractor initialization and our ensemble of classifiers. Finally, in Section~\ref{subsec:robot}, we consider two case studies for realistic deployment of our methods: outlier detection of unwanted objects in data and deployment on robot platforms.

\subsection{Comparison to State-of-the-art Methods}
\label{subsec:sota}
In Table \ref{table:multi-results}, we compare our method to two state-of-the-art approaches: the ResNet-50 classifier in \cite{raine2020multi} and the EfficientNet-B5 classifier reported in \cite{noman2021multi}\footnote{The source code from \cite{noman2021multi} was not made publicly available and  we re-implemented their method to the best of our ability.}. We also compare to two competitive baseline approaches: using CLIP \cite{radford2021learning} as a zero-shot classifier; and using SimCLR \cite{chen2020simple} to pre-train the feature extractor, and then train a linear classification head in a fully supervised manner on the patches from \cite{raine2020multi}.

As shown in Table~\ref{table:multi-results}, our approach outperforms all prior approaches and baseline methods for the per-class and overall F1 scores by a large margin in the case of multi-species seagrass classification.  Our method improves on the overall F1 score by 6.8\% and 7.9\% for the multi-species case when compared to \cite{noman2021multi} and \cite{raine2020multi} respectively.  Our method is also comparable to prior approaches in terms of inference time (\textasciitilde40 fps, as seen in Table~\ref{table:multi-results}).

Table~\ref{table:binary-results} shows that we also outperform \cite{noman2021multi} and \cite{raine2020multi} on the binary seagrass presence/absence problem, with an improvement of 7.0\% and 7.9\% respectively.

\subsection{Ablation Studies}
\label{subsec:ablations}

\subsubsection{Effect of Feature Extractor Initialization}
\label{subsubsec:feature}
We find that training our feature extractor using the SimCLR unsupervised contrastive pretext task \cite{chen2020simple} significantly improves the performance of our custom loss function for fine-tuning our architecture, with an improvement of 15.2\% and 8.1\% when compared to initialization with ImageNet and random weights respectively (Table \ref{table:features}).  The feature extractor trained on ImageNet does not transfer well to the seagrass classification task, whereas using the randomly initialized feature extractor captures spatial features and projects the images into a discriminative embedding space without the unhelpful bias learnt from the ImageNet pre-training task. 
\vspace*{-0.25cm}

\begin{table}[t]
\setlength{\tabcolsep}{4px}
\caption{SeaFeats Feature Extractor Initialization -- DeepSeagrass F1 Scores (Refer to Section~\ref{subsec:evaluationmetrics} for Metric Definitions)} 
\vspace*{-0.2cm}
\label{table:features}
\centering
\scriptsize
\begin{tabularx}{0.98\columnwidth}{@{}l>{\centering\arraybackslash}X>{\centering\arraybackslash}X>{\centering\arraybackslash}X>{\centering\arraybackslash}X>{\centering\arraybackslash}X}
\toprule
\textbf{Method} & \textbf{Back.} & \textbf{Fern.} & \textbf{Roun.} & \textbf{Strap.} & \textbf{Overall}\\
\midrule
SeaFeats: Random init & 86.48 & 84.54 & 86.10 & 82.16 & 85.05 \\
\arrayrulecolor{black!30}\midrule
SeaFeats: ImageNet init & 72.62 & 90.29 & 55.18 & 80.69 & 77.89 \\
\midrule
SeaFeats: SimCLR \cite{chen2020simple} init & \textbf{94.93} & \textbf{92.44} & \textbf{89.56} & \textbf{92.12} & \textbf{93.10} \\
\arrayrulecolor{black!100}\bottomrule 
\end{tabularx}
\vspace*{-0.2cm}
\end{table}

\subsubsection{Ensemble of Classifiers}
\label{subsubsec:weights}
\begin{figure}[t]
    \vspace*{0.15cm}
    \centering
    \includegraphics[width=\linewidth, clip, trim=5.7cm 3.9cm 6cm 4.9cm]{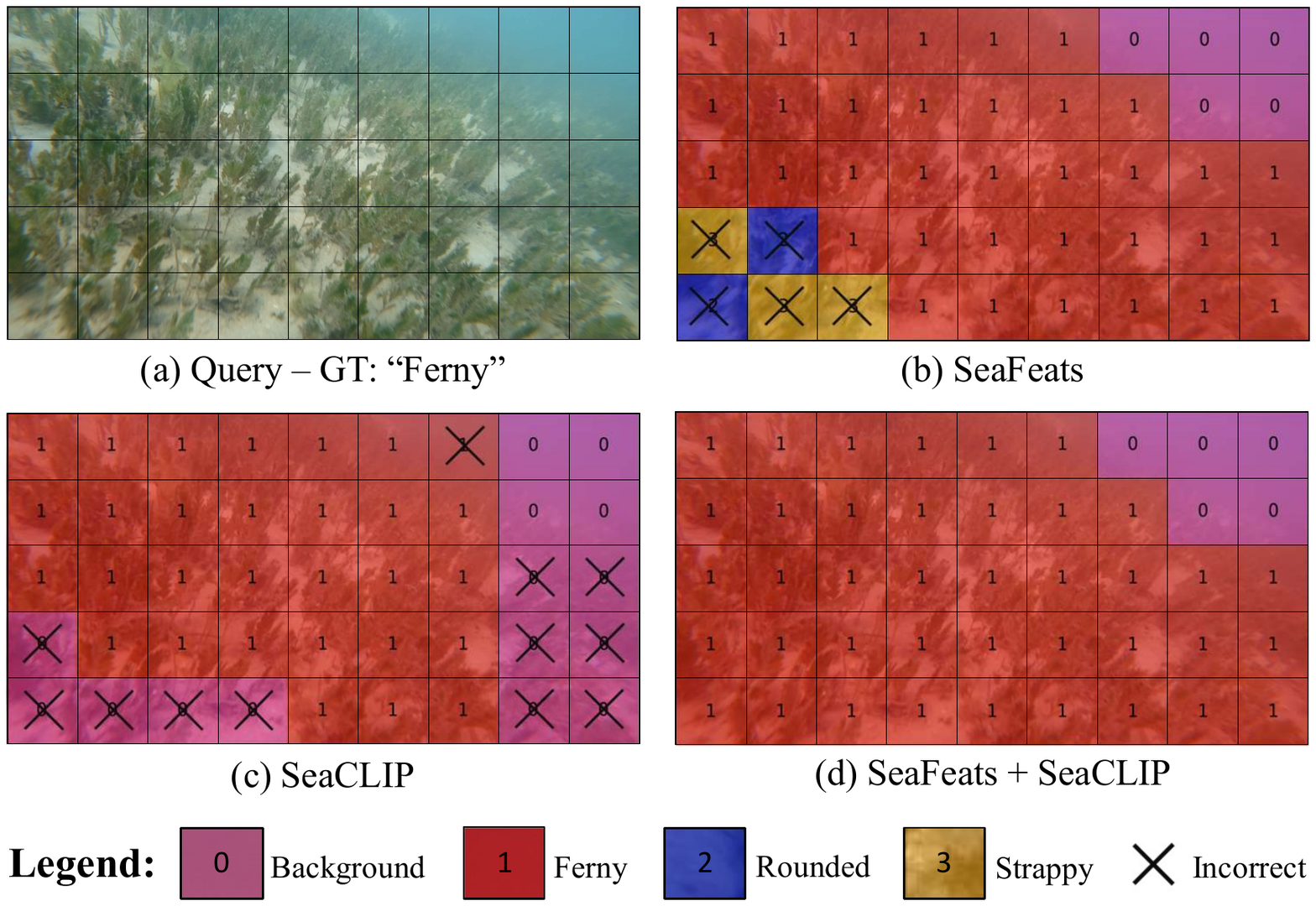}
    \vspace*{-0.3cm}
    \caption{For the query image shown in (a), our ensemble of classifiers combines the output of SeaFeats (b) and SeaCLIP (c) to give the combined softmax values in (d).  In (b), the blurry patches on the bottom left of the image have been incorrectly predicted as Strappy and Rounded seagrass. In (c), SeaCLIP has conservatively predicted these patches as background. Combining both models (d) prevents incorrect classification and the patches are correctly classified as Ferny seagrass.}
    \label{fig:ensemble}
    \vspace*{-0.4cm}
\end{figure}

We find that the performance of the feature similarity classifier (SeaFeats) is more robust when combined in an ensemble with our CLIP pseudo-labeling approach (SeaCLIP), with an improvement of 2.2\% and 7.5\% in the absolute F1 score as compared to only SeaFeats or only SeaCLIP, respectively (Table \ref{table:ensemble}).  Recall that as detailed in Section~\ref{subsec:met-ensemble}, the normalized output logits from each model are averaged at inference time to improve confidence for uncertain classifications. As seen in Fig.~\ref{fig:ensemble}, the combination of SeaCLIP with our generally higher-performing SeaFeats model results in overall improved performance due to the influence of SeaCLIP on visually degraded patches. We provide additional qualitative results in the Supplementary Material. 

\begin{table}[t]
\setlength{\tabcolsep}{4px}
\caption{Effect of Ensemble of Classifiers -- DeepSeagrass F1 Scores (See Section~\ref{subsec:evaluationmetrics} for Metric Definitions)} 
\vspace*{-0.2cm}
\label{table:ensemble}
\centering
\scriptsize
\begin{tabularx}{0.98\columnwidth}{@{}l@{\hspace{5.3em}}>{\centering\arraybackslash}X>{\centering\arraybackslash}X>{\centering\arraybackslash}X>{\centering\arraybackslash}X>{\centering\arraybackslash}X}
\toprule
\textbf{Method} & \textbf{Back.} & \textbf{Fern.} & \textbf{Roun.} & \textbf{Strap.} & \textbf{Overall}\\
\midrule
SeaFeats  & 94.93 & 92.44 & 89.56 & 92.12 & 93.10 \\
\arrayrulecolor{black!30}\midrule
SeaCLIP  & 86.14 & 87.82 & 87.91 & 91.21 & 87.84 \\
 \midrule
SeaFeats+SeaCLIP & \textbf{94.97} & \textbf{95.42} & \textbf{94.69} & \textbf{96.29} & \textbf{95.33}  \\
\arrayrulecolor{black!100}\bottomrule 
\end{tabularx}
\vspace*{-0.1cm}
\end{table}

\subsection{Case Studies for Robot Deployment}
\label{subsec:robot}

In this section, we present two case studies for real-world deployment of our method. First, we consider the `Global Wetlands' dataset which emulates the scenario where an ecologist collects a new set of images which contain some objects the ecologist is not interested in.  We then evaluate the generalization capability of our model to new platforms by applying our method on a survey transect collected by the FloatyBoat \cite{mou2022reconfigurable} autonomous surface vehicle.

In our first case study, we show the effectiveness of CLIP in a problem where the labels are even more limited: image-level labels are not available and all images contain multiple classes.  This setting mimics the scenario where an ecologist has collected a set of seagrass images, however the images may contain out-of-distribution classes which we want our classifier to separate from the seagrass and background patches (in this case, the `fish' class).  We leverage CLIP to predict which image patches belong to each class, acting as the supervisory signal for training our SeaCLIP classifier (refer to the Supplementary Material for details on the prompts we use). 

\begin{figure}
    \centering
    \includegraphics[width=0.95\linewidth, clip, trim=5.7cm 3.7cm 7.2cm 3.6cm]{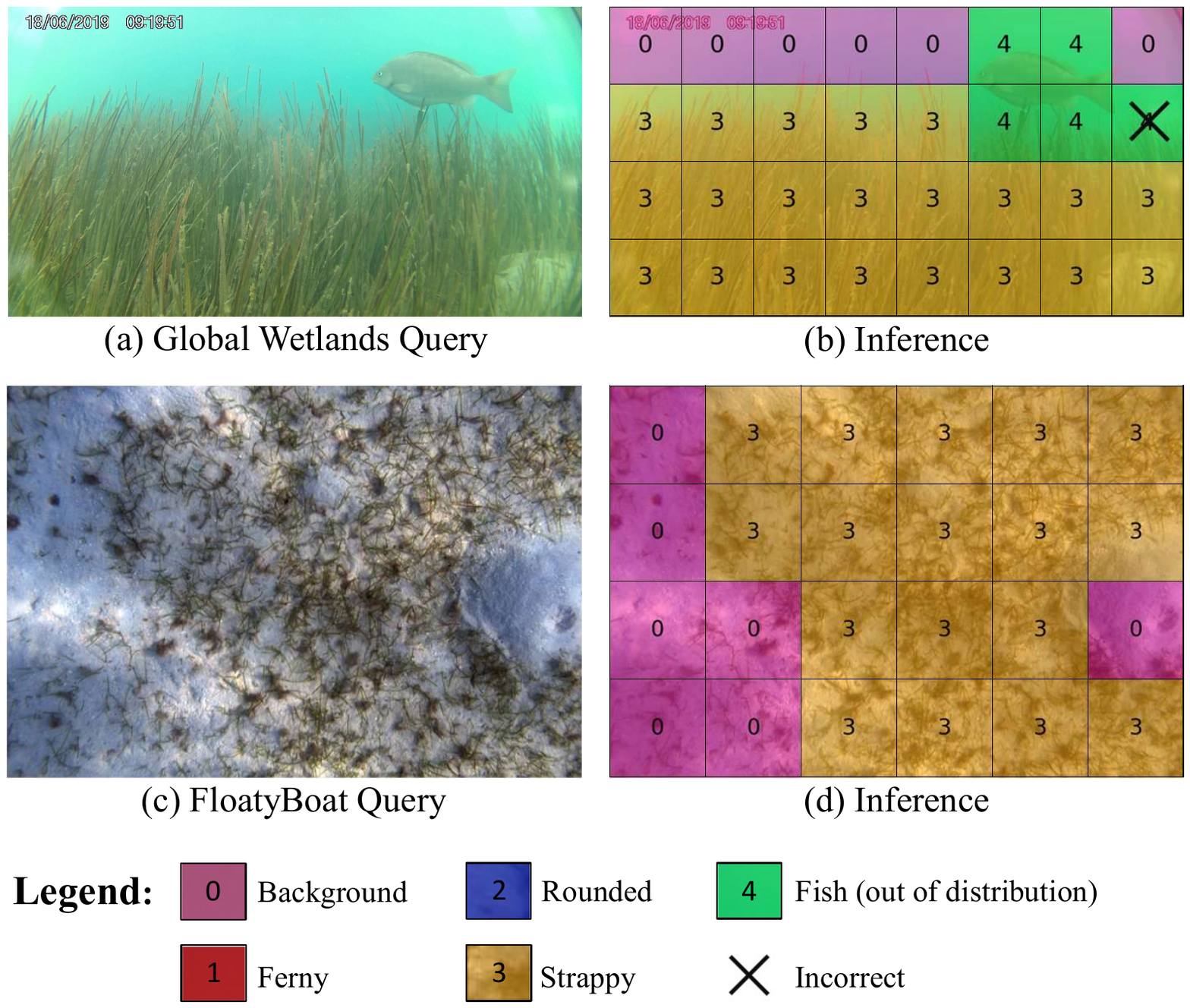}
    \caption{Qualitative evaluation of model generalization to other platforms. Sample test image from Global Wetlands (a) and inference from our SeaCLIP model in (b).  A transect image collected by top-down camera onboard FloatyBoat \cite{mou2022reconfigurable} is shown in (c), with corresponding inference from the fine-tuned SeaFeats in (d).}
    \label{fig:robot}
    \vspace*{-0.4cm}
\end{figure}

SeaCLIP effectively separates the seagrass, fish, and background classes, achieving 87.9\% F1 score with only CLIP as supervision during training (Table \ref{table:fish-results} and Fig.~\ref{fig:robot}).  It is also important to note that in this scenario, where the classes are broader and more visually distinct than in the multi-species seagrass case, the zero-shot CLIP model is able to achieve 85.8\% accuracy.  SeaCLIP further improves performance by 2.1\% compared to using CLIP and has the distinct advantage of performing inference \textasciitilde30 times faster than zero-shot CLIP (Table~\ref{table:multi-results}). 

We also evaluate the capacity of our DeepSeagrass-trained method to generalize across the different camera properties and visual conditions of the Global Wetlands dataset. Our ensemble method outperforms \cite{noman2021multi} and \cite{raine2020multi} for the seagrass presence/absence problem (``Binary Overall'' in Table~\ref{table:fish-results}) by 12.1\% and 28.9\% respectively for the absolute F1 score, and even outperforms SeaCLIP trained on the Global Wetlands dataset by 1.3\%, even though our ensemble was only trained on DeepSeagrass. 

\begin{table}[t]
\caption{Global Wetlands F1 Scores (See Section~\ref{subsec:evaluationmetrics} for Metric Definitions)} 
\vspace*{-0.15cm}
\label{table:fish-results}
\centering
\scriptsize
\setlength{\tabcolsep}{2pt}
\begin{tabular}{@{} lcccccc @{}}
 \toprule 
\multirow{2}{*}{\textbf{Method}} & \textbf{Training} & \textbf{Back-} & \multirow{2}{*}{\textbf{Fish}} & \textbf{Sea-} & \textbf{Multi-class} & \textbf{Binary} \\
 & \textbf{Dataset} & \textbf{ground} & & \textbf{grass} & \textbf{Overall} & \textbf{Overall} \\
\midrule
Zero-shot CLIP \cite{radford2021learning} & Global Wetlands & 84.55 & \textbf{60.73} & 90.54 & 85.77 & 88.01  \\
\arrayrulecolor{black!30}\midrule
SeaCLIP & Global Wetlands & 87.97 & 59.57 & 91.81 & \textbf{87.86} & 90.86 \\
\arrayrulecolor{black!100}\midrule
Raine \etal \cite{raine2020multi} & DeepSeagrass & 30.04 & n/a & 75.04 & n/a & 63.21 \\
\arrayrulecolor{black!30}\midrule
Noman \etal \cite{noman2021multi} & DeepSeagrass & 69.72 & n/a & 85.15 & n/a & 80.07 \\
\midrule
SeaFeats+SeaCLIP & DeepSeagrass  & \textbf{89.93} & n/a & \textbf{93.58} & n/a & \textbf{92.16} \\
\arrayrulecolor{black!100}\bottomrule 
\end{tabular}
\vspace*{-0.45cm}
\end{table}

In our second case study, we evaluate the performance of our method on images collected by an autonomous surface vehicle \cite{mou2022reconfigurable}. We fine-tune our method on only 10 background images and 10 seagrass images for 10 epochs.  We demonstrate that our model is able to generalize to a completely different platform and image characteristics, and is able to predict Strappy seagrass and background patches correctly in each frame, as seen in Fig.~\ref{fig:robot}. We note that we could not quantitatively evaluate our method on this dataset as ground truth labels are not available, but both Fig.~\ref{fig:robot} and the accompanying video demonstrate qualitatively that our method performs well in this setting.

\vspace*{-0.025cm}
\section{Conclusions}
\label{sec:conclusions}
\vspace*{-0.05cm}

In this work, we proposed a weakly supervised method for coarse multi-species segmentation of seagrass, which requires only image-level annotations. Our approach performs seagrass detection and classification at the patch-level during inference using two complementary methods, SeaFeats and SeaCLIP. %
Our proposed ensemble (SeaFeats+SeaCLIP) achieves a class-weighted F1 score of 95.3\% on DeepSeagrass, outperforming the prior state-of-the-art by 6.8\%.

We demonstrated the relevance of pre-trained large vision-language models to the field of ecology and marine surveys, and showed that CLIP provides an effective supervisory signal for weakly supervised training. Our method could be adapted to other applications such as mapping weeds in precision agriculture or for coarse segmentation of remotely sensed imagery for land and vegetation cover analysis.

Our method has been designed with a robotics use-case in mind. The real-time inference will create exciting new research opportunities in adaptive navigation and surveys. For instance, our method could facilitate the survey of high-density seagrass areas or areas with rare seagrass species at a higher level of detail compared to less interesting areas such as sand patches. This capability is particularly critical given the limited prior knowledge about survey areas, as remote observation methods do not provide sufficient detail to guide underwater or surface autonomous vehicles.
Other future works could involve pixel-wise semantic segmentation of multi-species seagrass images, seagrass density estimation, and blue carbon stock estimation from our model inferences.

%

\begin{comment}
***
CLIP performance degrades when images get smaller.  This approach uses high resolution imagery, however if high resolution imagery is not available then the ability of CLIP to generate accurate pseudo-labels is likely to decrease.

Future work:
-seagrass density estimation?
-seagrass segmentation
-use of domain adaptation for better transfer to underwater vehicle?
-use to estimate blue carbon stock?
***
\end{comment}
%
\vspace*{-0.25cm}
{\small
\section*{Acknowledgments}
\vspace*{-0.05cm}
This work was done in collaboration between QUT and CSIRO Data61.  S.R., F.M., and T.F.~acknowledge continued support from the Queensland University of Technology (QUT) through the Centre for Robotics. T.F.~acknowledges funding from ARC Laureate Fellowship FL210100156 and Intel Labs via grant RV3.290.Fischer.

\bibliographystyle{ieee_fullname}
\bibliography{Bibliography}
}

\end{document}

\title{Image Labels Are All You Need for Coarse Seagrass Segmentation \\ 
\vspace{\baselineskip}
\large{Supplementary Material}}

\author{Scarlett Raine$^{1,2}$, Ross Marchant$^{3}$, Brano Kusy$^{2}$, Frederic Maire$^{1}$ and Tobias Fischer$^{1}$ \\
\\
\footnotesize{$^{1}$QUT Centre for Robotics, Queensland University of Technology, Australia} {\tt \footnotesize \emph{\{sg.raine, f.maire, tobias.fischer\}@qut.edu.au}} \\
\footnotesize{$^{2}$CSIRO Data61, Australia} {\tt \emph{\{scarlett.raine, brano.kusy\}@csiro.au}} \\
\footnotesize{$^{3}$Image Analytics, Australia} {\tt \emph{ross.g.marchant@gmail.com}}%
}

\renewcommand{\figurename}{Supplementary Figure}

\maketitle
\section*{Overview}

This is Supplementary Material for the paper, `Image Labels Are All You Need for Coarse Seagrass Segmentation'. We further explore the performance of our introduced ensemble of classifiers, SeaFeats and SeaCLIP. Section \ref{sec:results} supplements the results presented in the main paper with additional qualitative examples of failure cases and analysis for each case.  We also provide additional implementation details in Section \ref{sec:implementation}.

\section{Additional Qualitative Results}
\label{sec:results}

In Section 5.2.2 of our main paper, we present the output results when combining our classifiers, SeaFeats and SeaCLIP, in an ensemble.  This combination exhibits superior performance than using either classifier individually, because the generally higher-performing SeaFeats model benefits from the conservative predictions of SeaCLIP to result in more robust performance overall.  In this section, we further analyze the effect of combining SeaFeats and SeaCLIP, and we focus on failure cases which result in incorrect predictions.  For each example, we also compare our qualitative predictions to the outputs of our re-implementation of the EfficientNet-B5 approach presented in \cite{noman2021multi}.

\begin{figure*}[t]
\centering
\includegraphics[width=\linewidth,clip, trim=0cm 5cm 0cm 2cm]{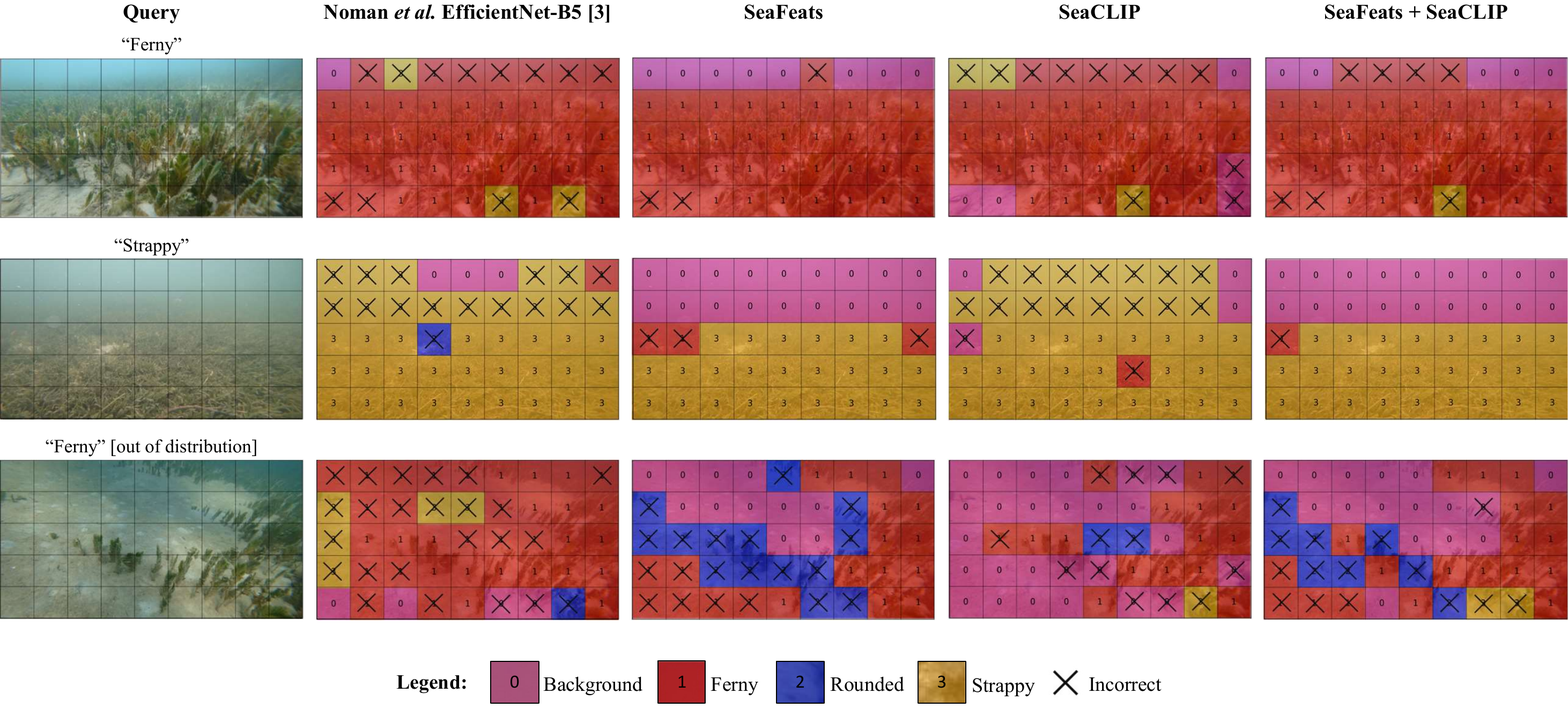}
\caption{Example cases where both the prior approach \cite{noman2021multi} and our proposed classifiers fail. Top row: Many failure cases occur around the edge of images due to camera distortion, blurring, or darker regions in these parts of the images. Middle row: Some images may be visually degraded due to turbidity, lighting, or algae, resulting in failure cases. Bottom row: When models are deployed on images from outside the distribution of the training data, the species of seagrass are more likely to be incorrectly assigned.}
\label{fig:eg1}
\end{figure*}

There are a range of factors which may result in a failure case: edge patches which are subject to camera distortion, resulting in blurry, darkened or warped image patches (Supp.~Fig.~\ref{fig:eg1}, first row); significant difference in scale or resolution of inference imagery; visually degraded images due to turbidity, lighting or algae (Supp.~Fig.~\ref{fig:eg1}, second row); inference on images which contain out-of-distribution seagrass species or seagrasses at a different stage of growth than seen in the training set (Supp.~Fig.~\ref{fig:eg1}, third row); or presence of unknown objects in inference images.  These factors are largely caused by limitations of the training data -- a larger, more varied dataset which encompasses a wider range of conditions, visual characteristics, seagrass appearance changes and image qualities would improve the ability of the models to generalize to previously unseen conditions.

Supp.~Fig.~\ref{fig:eg1} (first row) demonstrates the impact of camera distortion and blur for image patches at the edge of images.  SeaFeats and SeaCLIP are more likely to incorrectly classify these patches than the clear image patches at the center of the image. SeaFeats classifies the majority of the patches correctly, however SeaCLIP is uncertain about multiple patches and misclassifies some edge patches as the `Strappy' species of seagrass (yellow) and others as `Background' (pink). This example also demonstrates a failure case for the ensemble of classifiers, such that the correct predictions from one model (SeaFeats) are overridden by the incorrect predictions of the other (SeaCLIP).  Although there are examples when this occurs, in general the ensemble of classifiers results in improved and more robust performance, as demonstrated in the results section of the main paper.

Supp.~Fig.~\ref{fig:eg1} (second row) illustrates that images degraded due to turbidity, lighting and/or algae may result in incorrectly classified patches.  In this example, the water column has high levels of turbidity, resulting in an image with a foggy appearance.  Although the majority of patches in the image are correctly classified (`Strappy' class in yellow), all models incorrectly classify a few patches. This type of failure case could be mitigated by increasing the range of visual characteristics in the training dataset.

Supp.~Fig.~\ref{fig:eg1} (third row) demonstrates that all models incorrectly classify image patches where the seagrass has a different visual appearance as compared to the training dataset. Here, the \textit{Halophila spinulosa} seagrass is not as dense as in the training dataset (particularly in the center of the image), and the seagrass is at an earlier stage of growth. The distribution of training examples seen by the model needs to encompass all stages of seagrass growth and other factors including presence/absence of algae, season and weather conditions to ensure that models can effectively transfer to a variety of inference images.

These failure cases demonstrate how the availability and variation of training data impacts on model performance.  Future work could include human-in-the-loop training or bootstrapping to iteratively update models based on expert verification or correction of model inferences during deployment.

\section{Additional Implementation Details}
\label{sec:implementation}

\subsection{SeaCLIP}

\begin{figure}[t]
\centering
\includegraphics[width=\linewidth, clip, trim=1.9cm 11.75cm 8.1cm 2.2cm]{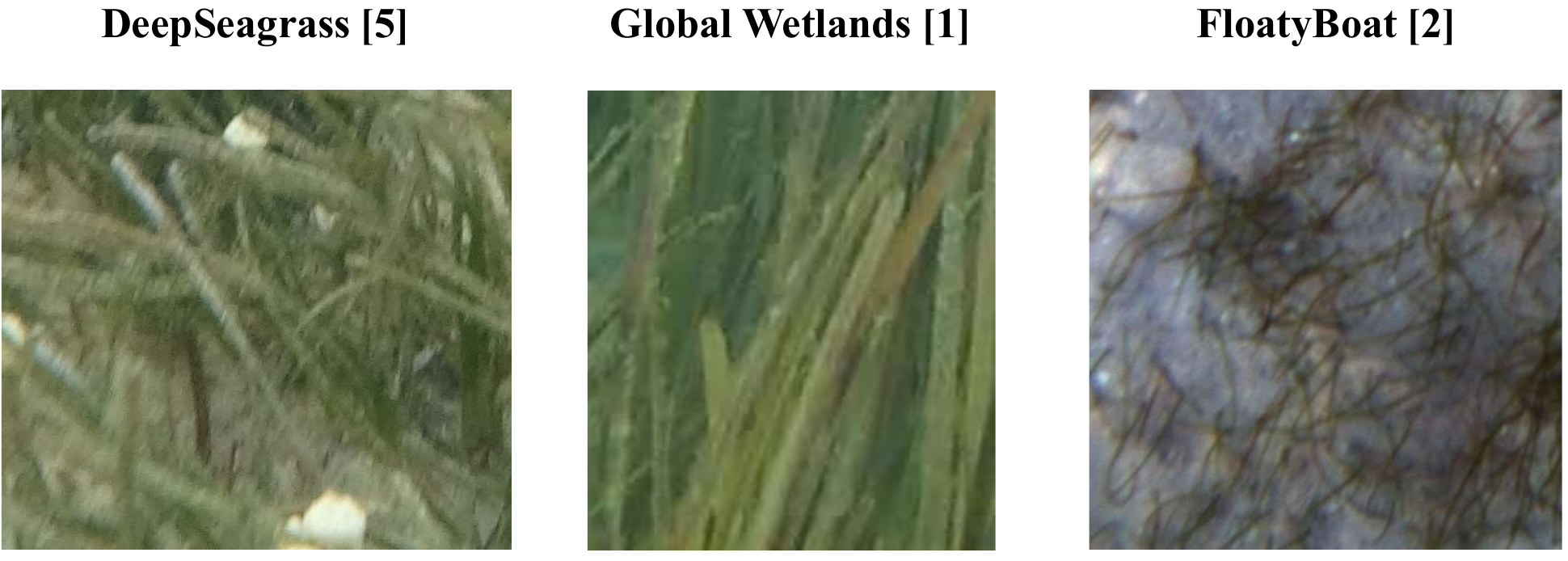}
\caption{Example patches from each of the datasets: DeepSeagrass \cite{raine2020multi} (left), Global Wetlands \cite{ditria2021annotated} (center) and FloatyBoat \cite{mou2022reconfigurable} (right).  The patch size between the datasets differ, but are selected to maintain a similar scale in terms of seagrass appearance.}
\label{fig:patch}
\end{figure}

We train SeaCLIP on image patches pseudo-labeled by the CLIP large language model \cite{radford2021learning}.  We use patches of size 520x578 pixels (Supp.~Fig.~\ref{fig:patch}), as per the DeepSeagrass dataset \cite{raine2020multi}.  The query phrases used for obtaining the binary pseudo-labels are: 
\begin{itemize}
    \item `Background': ``a photo of sand'', ``a photo of water'', ``a photo of sand or water'', ``a blurry photo of water'', ``a blurry photo of sand''; and
    \item `Seagrass': ``a blurry photo of seagrass'', ``a photo containing some seagrass'', ``a photo of underwater plants'', ``a photo of underwater grass'', ``a photo of green, grass-like leaves underwater'', ``a photo of seagrass''.
\end{itemize}
We assign the image-level seagrass species (i.e.~`Ferny', `Rounded' or `Strappy') to the patches pseudo-labeled by CLIP as `Seagrass' at training time.  

\subsection{Inference on Global Wetlands Dataset}
We use the Global Wetlands dataset \cite{ditria2021annotated} to evaluate the ability of our proposed models to generalize to unseen data and to assess the performance of the SeaCLIP model.  We process the Global Wetlands dataset by splitting images into 50 (10x5) patches, resulting in a patch size of 192x216 pixels.  We selected this grid size to maintain a similar scale for the seagrass within each image patch as for the DeepSeagrass patches (as seen in Supp.~Fig.~\ref{fig:patch}).  

When evaluating the performance of CLIP as a zero-shot classifier on this dataset, we use the following prompts:
\begin{itemize}
    \item `Background': ``a photo of sand'', ``a photo of blue water'', ``a photo of murky, green water'', ``a photo of sand or water'', ``a blurry photo of water'', ``a blurry photo of sand''; 
    \item `Seagrass': ``a blurry photo of seagrass'', ``a photo containing some seagrass'',
     ``a photo of underwater plants'', ``a photo of underwater grass'', ``a photo of green, grass-like leaves underwater'', ``a photo of seagrass''; and
    \item `Fish': ``a photo of fish'', ``a close-up photo of fish'', ``a blurry photo of fish'', ``a photo containing part of a fish'', ``a photo of fish scales''.
\end{itemize}

When training SeaCLIP on Global Wetlands, the CLIP-generated pseudo-labels were created using the same prompts as above. 

\subsection{Fine-tuning for FloatyBoat Dataset}

For evaluation of model generalization to the FloatyBoat \cite{mou2022reconfigurable} autonomous surface vehicle dataset, we take the SeaFeats model trained on DeepSeagrass and fine-tune it on 10 background images and 10 seagrass images for 10 epochs. We use a range of data augmentations to improve the ability of the model to generalize to the different camera characteristics and imagery viewpoint: we randomly apply augmentations which vary the color channels, linear contrast, Gaussian blur, brightness, hue and saturation.  We additionally apply geometric augmentations including x and y scaling, and left/right flipping. 

At inference time, our approach assumes a 6x4 grid for each FloatyBoat image, resulting in patches which are 468x455 pixels.  Similarly to the Global Wetlands dataset, this grid size is selected so that the seagrass appears at a similar scale within each patch as the DeepSeagrass dataset (Supp.~Fig.~\ref{fig:patch}). 

{\small
\bibliographystyle{ieee_fullname}
\bibliography{Bibliography}
}